\documentclass[times, review, 10pt]{elsarticle}
\usepackage{amssymb}
\usepackage{amsmath}
\usepackage{amsthm}

\usepackage{multirow}
\usepackage{diagbox}
\usepackage{enumitem}

\usepackage[linesnumbered,ruled]{algorithm2e}

\usepackage{tikz}
\usetikzlibrary{positioning}
\usetikzlibrary{shapes}
\usetikzlibrary{arrows}

\usepackage{caption}
\usepackage{subcaption}

\newtheorem{definition}{Definition}
\newtheorem{property}{Property}
\newtheorem{example}{Example}
\renewcommand{\Pr}{\mathbb{P}}

\journal{Pattern Recognition}

\begin{document}

\begin{frontmatter}



\title{Introducing the O-Value: A Universal Standardization for Confusion-Matrix-Based Classification Performance Metrics}

%\author{Ningsheng Zhao\fnref{aff1,cor1}}
\author{Ningsheng Zhao\fnref{aff1}}
%\ead{ningsheng.zhao@mail.concordia.ca}
%\fntext[cor1]{currently at Harvard School of Dental Medicine, Boston, MA, USA 02115}

%\author{Trang Bui\fnref{aff2,cor2}\corref{cor3}}
\author{Trang Bui\fnref{aff2}\corref{cor3}}
\ead{tqtbui@uwaterloo.ca}
\cortext[cor3]{Corresponding Author}
%\fntext[cor2]{currently at University of Rochester, Rochester, NY, USA 14642}

\author[aff1]{Jia Yuan Yu}

\author[aff1,aff3]{Krzysztof Dzieciolowski}

\affiliation[aff1]{organization={Concordia University},
             addressline={1455 Blvd. De Maisonneuve Ouest},
             city={Montreal},
             postcode={H3G 1M8},
             state={Quebec},
             country={Canada}}
\affiliation[aff2]{organization={University of Waterloo},
             addressline={200 University Ave West},
             city={Waterloo},
             postcode={N2L 3G1},
             state={Ontario},
             country={Canada}}
\affiliation[aff3]{organization={Daesys Inc.},
             addressline={3 Pl. Monseigneur Charbonneau suite 400},
             city={Montreal},
             postcode={H3B 2E3},
             state={Quebec},
             country={Canada}}


%% Abstract
\begin{abstract}
Many classification performance metrics exist, each suited to a specific application. However, these metrics often differ in scale and can exhibit varying sensitivity to class imbalance rates in the test set. As a result, it is difficult to use the nominal values of these metrics to evaluate, compare and monitor classification performances, especially when imbalance rates vary. To address this problem, we introduce the outperformance standardization (OPS) function, a universal standardization method for confusion-matrix-based classification performance (CMBCP) metrics. It maps any given metric to a common scale of $[0,1]$, while providing a clear and consistent interpretation. Specifically, the resulting OPS value (o-value) represents the percentile rank of the observed classification performance within a reference distribution of possible performances. This unified framework enables meaningful comparison and monitoring of classification performance across test sets with differing imbalance rates. We illustrate how o-values can be applied to a variety of commonly used classification performance metrics and demonstrate the utility and robustness of our method through experiments on real-world datasets spanning multiple classification applications.
\end{abstract}


%    \item Class imbalance can influence the range and interpretation of the nominal values of many classification performance metrics. There are no common criteria for evaluating the nominal values of these metrics.  
%    \item We formalize a mathematical framework for confusion-matrix-based classification performance (CMBCP) metrics. We categorize them into labeling and scoring metrics and derive their respective properties. 
%    \item We propose a universal standardization method called the outperformance standardization (OPS) function. The function can be applied to any CMBCP metrics, and the resulting OPS values (o-values) will be scale-free and have a clear interpretation. 
%    \item Our proposed o-values can be used to evaluate, compare, and monitor performances across datasets with varying imbalance rates.
%    \item Via experiments on real-world datasets, we illustrate the robust performance and utility of our method when applied to different performance metrics for different classification applications. 

%% Keywords

\end{frontmatter}



%-------------------------------------
\section{Introduction} \label{sec:intro}
%-------------------------------------

Classification is an important machine learning task broad applications in finance, healthcare, business, and beyond. In practice, classifiers are trained on large-scale datasets \cite{l2017machine}, often requiring substantial computational time and effort. Therefore, interpretable and reliable performance metrics are essential for evaluating and tracking performance so that developers are alerted when retraining is necessary \cite{domingos2012few,amershi2015modeltracker}. 

There have been many performance metrics proposed in the literature to evaluate classifiers, such as f1\_score \cite{christen2023review}, Matthews Correlation Coefficient (MCC) \cite{matthews1975comparison}, Receiver Operating Characteristic (ROC) curve \cite{fawcett2006introduction}, Precision-Recall curve (PRC) \cite{powers2011evaluation}, lift curve \cite{tuffery2011data}, and so on. Because classification tasks vary widely in their objectives, different applications favor different metrics \cite{powers2011evaluation,japkowicz2011evaluating,chicco2021matthews}. While the ROC is preferred for balanced data, the PRC is more favorable in imbalanced datasets \cite{saito2015precision}. Other authors recommend using the MCC instead of the ROC for reliable classification evaluation \cite{chicco2021matthews}. Meanwhile, each performance metric has a different interpretation, and depending on the test set, they may have different ranges of values. As a result, direct evaluation, comparison, or monitoring based on nominal metric values is often challenging, particularly when the test set changes. In other words, there are no universal criteria to evaluate classifiers and detect model drift.

These issues are further amplified in applications such as fraud detection, disease diagnosis, and relevance ranking, where classifiers are trained and evaluated on imbalanced data. Many methods have been proposed to train and improve the classification performance under class imbalance \cite{ferri2009experimental,tarekegn2021review}. Examples include Synthetic Minority Over-sampling Technique (SMOTE) \cite{chawla2002smote,maldonado2022fw}, data augmentation \cite{suh2022discriminative,ding2025improving}, cost-sensitive learning \cite{sun2007cost,koyejo2014consistent,lazaro2023neural}, and so on. In terms of performance metrics, multiple proposals have been proposed to overcome the problem of imbalance, such as the Class Balance Accuracy \cite{brodersen2010balanced}, Index of Balance Accuracy \cite{garcia2009index}, relevance-based performance metric \cite{branco2017relevance}, normalized PRC \cite{powers2011evaluation}, or normalized precision \cite{daskalaki2006evaluation}. Sensitivity of metrics to class imbalance is also studied in \cite{gu2009evaluation,luque2019impact}. Nevertheless, these methods target a specific interest and/or application and do not provide a universal interpretation for evaluation.

In this article, we aim to provide a standardization method that practitioners can apply to their preferred confusion-matrix-based classification performance (CMBCP) metric. The resulting standardized values will have a universal and clear interpretation, which enables consistent evaluation, comparison, and monitoring of classification performances across different test sets with possibly different class imbalance rates. Our article expands on the preliminary work of \cite{zhao2022classifier}, where the key idea is to evaluate classification performance by comparing it to all possible performances given the test set. In particular, we provide a detailed mathematical formulation, justifications, explanations, and experiments to demonstrate how such an idea can be implemented. 

The rest of the paper is organized as follows. In Section~\ref{sec:metrics}, we present the mathematical settings, introduce two types of CMBCP metrics, and discuss their issues. We introduce the outperformance standardization (OPS) function, a universal standardization method for CMBCP metrics, in Section~\ref{sec:method}. Section~\ref{sec:experiment} demonstrates the use of our proposed outperformance standardization method on real-world datasets. Finally, summaries of our contributions and directions for future research are provided in Section~\ref{sec:conclusion}.  

%-------------------------------------
\section{Classifier Performance Metrics} \label{sec:metrics}
%-------------------------------------

\subsection{Preliminary}  \label{sec:prelim}

Consider a binary classification problem, where the goal is to predict label $Y \in \{0, 1\}$ based on features $X = (X_1, ..., X_d) \in \mathcal{X} \subset \mathbb{R}^d$ given the training set $\mathcal{D}_{\mathrm{train}} = \{(\mathbf{x}_i, y_i)\}_{i=1}^m$, where $y_i$ is the true class label of the $i$th instance (observation) with features $\mathbf{x}_i$. A classifier $\hat{f}$ can be obtained by training a classification model $f$ on the data set $\mathcal{D}_{\mathrm{train}}$. Depending on the classification model $f$, given an input $\mathbf{x} \in \mathcal{X}$, the classifier $\hat{f}$ can output either a binary label $\hat{f}(\mathbf{x}) \in \{0,1\}$, or a score $\hat{f}(\mathbf{x}) \in [0,1]$. Note that for some classifiers such as logistic regression, the score $\hat{f}(\mathbf{x})$ is exactly the predicted probability that the instance associated with input $\mathbf{x}$ has class membership 1 instead of 0. In other cases, such as naive Bayes, support vector machine, or any tree-based classifier, these scores are only related, but not exactly equal to, the predicted probability \citep{zadrozny2002transforming,kull2017beyond}. 

When the output $\hat{f}(\mathbf{x})$ is a binary label, the predicted class membership $\hat{y}$ is equal to the classifier's output $\hat{f}(\mathbf{x})$. On the other hand, when $\hat{f}(\mathbf{x})$ is a score, a threshold $t$ must be applied to determine the predicted class membership
\begin{equation}
    \hat{y} = \begin{cases}
        1 & \text{if } \hat{f}(\mathbf{x}) \ge t \\
        0 & \text{otherwise} 
    \end{cases}. \label{eq:thresholding}
\end{equation}

To evaluate classifier $\hat{f}$, a test set $\mathcal{D}_{\mathrm{test}} = \{(\mathbf{x}_i, y_i)\}_{i=1}^n$ of size $n$ is used. Specifically, the classifier's outputs $\{\hat{f}(\mathbf{x}_i)\}_{i=1}^n$ are compared to the true labels $\{y_i\}_{i=1}^n$, and the results are summarized using some performance metrics. Many performance metrics are defined based on the \textit{confusion matrix}. 

\subsection{Confusion Matrix} \label{sec:confusion-matrix}

For each instance $i$ in the test set $\mathcal{D}_{\mathrm{test}}$, there are four possible combinations of the actual label $y_i$ and the predicted label $\hat{y}_i$: either $(1, 1), (0, 1), (1, 0)$, or $(0, 0)$. The confusion matrix is a $2\times 2$ table that counts the number of instances in the test set that fall into each of these four value combinations. 
\begin{table}[htbp!]
    \centering
    \begin{tabular}{|c|cc|}
        \hline
        \backslashbox{Predicted label}{True label} & 1 & 0 \\
        \hline
        1 & $n_1$ & $n_2$ \\  
        0 & $n_3$ & $n_4$ \\
        \hline
    \end{tabular}
    \caption{A confusion matrix}
    \label{tab:confusion-matrix}
\end{table}

As illustrated in Table~\ref{tab:confusion-matrix}, the confusion matrix consists of four elements $\{n_1, n_2, n_3, n_4\}$, where $n_1 = \sum_{i=1}^n \mathbb{I}(y_i = 1, \hat{y}_i = 1)$ counts the \textit{true positives} (TP); $n_2 = \sum_{i=1}^n \mathbb{I}(y_i = 0, \hat{y}_i = 1)$ counts the \textit{false positive} (FP); $n_3 = \sum_{i=1}^n \mathbb{I}(y_i = 1, \hat{y}_i = 0)$ counts the \textit{false negative} (FN); and $n_4 = \sum_{i=1}^n \mathbb{I}(y_i = 0, \hat{y}_i = 0)$ counts the \textit{true negative} (TN). It is easy to see that $n_1+n_2+n_3+n_4 = n$. From the confusion matrix, we can calculate the \textit{Type-I error} $\alpha$, \textit{Type-II error} $\beta$ and \textit{prevalence rate} $\pi$ (or positive class ratio):
\begin{equation}
    \alpha = \frac{n_2}{n_2+n_4}, \hspace{5mm} \beta = \frac{n_3}{n_1+n_3}, \hspace{5mm} \text{and} \hspace{5mm} \pi = \frac{n_1+n_3}{n}. \label{eq:alpha-beta-pi}
\end{equation}
We can see that the prevalence rate reflects class imbalance in the test set. When $\pi \ne 0.5$, there are fewer instances in one class compared to the other class, and therefore, the test set is imbalanced. 

%\begin{lemma}
%    For a confusion matrix as given in Table~\ref{tab:confusion-matrix}, $\{n, \pi, \alpha, \beta\}$ is a re-parametrization of $\{n_1, n_2, n_3, n_4\}$ if $\pi \in (0,1)$ (i.e., if $\alpha$ and $\beta$ are well-defined). \label{lem:reparametrization}
%\end{lemma}

%\begin{proof}
%    We can write $\{n_1, n_2, n_3, n_4\}$ as functions of $\{n, \pi, \alpha, \beta\}$, i.e.
%    \[n_1 = n\pi(1-\beta), \hspace{5mm} n_2 = n\pi\beta, \hspace{5mm} n_3 = n(1-\pi)\alpha, \hspace{5mm} \text{and} \hspace{5mm} n(1-\pi)(1-\alpha).\]
%    The Jacobian for this transformation is
%    \[\left[\begin{array}{cccc}
%    \pi(1-\beta) & n(1-\beta) & 0 & -n\pi \\
%    \pi\beta & n\beta & 0 & n\pi \\
%    (1-\pi)\alpha & -n\alpha & n(1-\pi) & 0 \\
%    (1-\pi)(1-\alpha) & -n(1-\alpha) & -n(1-\pi) & 0 \\
%    \end{array}\right].\]
%    The determinant of the Jacobian is $n^3\pi(1-\pi)$, which is non-zero when $\pi \ne 0$ and $\pi \ne 1$. In that case, the Jacobian is nonsingular, and $\{n, \pi, \alpha, \beta\}$ is a re-parametrization of $\{n_1, n_2, n_3, n_4\}$. 
%\end{proof}

It can be easily shown that when $\pi \in (0,1)$, $\{n,\pi,\alpha,\beta\}$ is a re-parametrization of $\{n_1,n_2,n_3,n_4\}$. That is, the information of a confusion matrix can be completely represented by the sample size $n \in \mathbb{N}$, the prevalence rate $\pi \in (0,1)$, the Type-I error rate $\alpha \in [0,1]$ and the Type-II error rate $\beta\in [0,1]$. See Lemma 1 of the Supplementary Material for a proof of this result.

Compared to $\{n_1, n_2, n_3, n_4\}$, the $\{n, \pi, \alpha, \beta\}$ highlights the different sources of variation in classification results: while $n$ and $\pi$ come from the test set, $\alpha$ and $\beta$ encode the two different types of classification errors. It is helpful to note that $\alpha = \text{FPR}$, while Type-II error $\beta = 1 - \text{TPR}$, where FPR is the false positive rate and TPR is the true positive rate. Thus, the $\alpha$-$\beta$ space has a one-to-one correspondence to the ROC space (i.e., the FPR-TPR space). 

As mentioned in Section \ref{sec:prelim}, many classification performance metrics are calculated based on the confusion matrix. In this paper, we focus on such confusion-matrix-based classification performance (CMBCP) metrics. Moreover, we classify them into two categories: \textit{labeling metrics} and \textit{scoring metrics}.

\subsection{Labeling Metrics} \label{sec:labeling-metrics}

Labeling metrics $\mathbf{M}_L$ are computed from a \textit{single} confusion matrix $C$. Examples of labeling metrics are \textit{recall}, \textit{precision}, and \textit{f1\_score} \citep{powers2011evaluation}, and the \textit{Matthews correlation coefficient} (MCC) \citep{chicco2021matthews}. Table~\ref{tab:labeling-metrics} provides the formulas of these metrics based on two parametrizations of the confusion matrix. The $\{n,\pi,\alpha,\beta\}$ parametrization reveals that these labeling metrics do not depend on the size of the test set $n$. However, metrics such as precision, f1\_score, and MCC \textit{do} depend on the prevalence rate $\pi$. 

\begin{table}[htbp!]
    \centering
    \begin{tabular}{|c|c|c|}
    \hline
        \textbf{Metrics}  & \textbf{In terms of $\{n_1, n_2, n_3, n_4\}$} & \textbf{In terms of $\{n, \pi, \alpha, \beta\}$}   \\
    \hline
        Recall & $\frac{n_1}{n_1+n_3}$ & $1-\beta$ \\
        Precision & $\frac{n_1}{n_1+n_2}$ & $\frac{\pi (1-\beta)}{\pi (1-\beta)+(1-\pi)\alpha}$ \\
        f1\_score & $\frac{2n_1}{2n_1+n_2+n_3}$  & $\frac{2 \pi (1-\beta)}{\pi(2-\beta)+(1-\pi)\alpha}$ \\
        MCC & $\frac{n_1  n_4 - n_2  n_3}{\sqrt{(n_1+n_2)(n_1+n_3)(n_4+n_2)(n_4+n_3)}}$ & $\frac{1-\alpha-\beta}{\sqrt{(1-\alpha+\frac{\pi}{1-\pi}\beta)(1-\beta+\frac{\pi}{1-\pi}\alpha)}}$  \\
    \hline
    \end{tabular}
    \caption{Formulas of some labeling metrics in terms of two parameterizations of the confusion matrix.} 
    \label{tab:labeling-metrics}
\end{table}

\subsection{Scoring Metrics} \label{sec:scoring-metrics}

When predicting labels of an instance with features $\mathbf{x}$, instead of directly outputting a binary label $\hat{y} \in \{0,1\}$, many machine learning classifiers such as neural networks or tree-based classifiers output a numeric score $\hat{f}(\mathbf{x}) \in [0,1]$. The label is then predicted by specifying a threshold $t \in (0,1)$ and applying it to Equation~\eqref{eq:thresholding}. Given the true labels $\{y_i\}_{i=1}^n$ and classifier output $\{\hat{f}(\mathbf{x}_i)\}_{i=1}^n$, the threshold $t$ uniquely defines a confusion matrix $C$. Although the threshold $t = 0.5$ is often used in practice, there is no stringent criterion to select such a threshold. This is due to two main reasons. First, the prediction score $\hat{f}(\mathbf{x})$ is not well-calibrated, in the sense that it does not represent the correct prediction likelihood despite being often interpreted as the predicted probability \citep{niculescu2005predicting, guo2017calibration, naeini2015obtaining}. Second, the threshold should be chosen based on the specific goal or application. For example, in marketing, it can be harmful to recommend an unwanted product, so a high threshold is usually used to reduce Type-I errors. In fraud detection, however, it's often preferable to flag a legitimate transaction than to miss a fraudulent one; therefore, a lower threshold is typically used to reduce Type-II errors.

This motivates the use of scoring metrics that summarize the overall performance of a classifier under all possible thresholds. Essentially, these metrics (curves) visualize certain trade-offs when different thresholds are applied. For example, the \textit{Receiver Operating Characteristics} (ROC) curve \citep{fawcett2006introduction} visualizes the trade-off between a low Type-I error rate $\alpha$ and a high recall $1-\beta$, while the \textit{Precision-Recall curves} (PRC) \citep{powers2011evaluation} visualize the trade-off between a high precision and a high recall. Table~\ref{tab:scoring-metrics} summarizes the formulas of popular scoring metrics curves. We again observe that the formulas of the x- and y-coordinates of these curves depend on $\pi$ and not on $n$. 

\begin{table}[htbp!]
    \centering
    \begin{tabular}{|c|l|l|c|}
    \hline
        \textbf{Curve} & \multicolumn{1}{|c|}{\textbf{x-coordinate}} & \multicolumn{1}{|c|}{\textbf{y-coordinate}} & \textbf{ideal $AUC$} \\
    \hline
        ROC & Type-I error: $\alpha$ & recall: $1-\beta$ & 1 \\
        PRC & recall: $1-\beta$ & precision: $\frac{\pi (1-\beta)}{\pi (1-\beta)+(1-\pi)\alpha}$ & 1 \\
        Lift Curve & percentage\footnotemark: $\pi (1-\beta)+(1-\pi)\alpha$ & lift: $\frac{(1-\beta)}{\pi (1-\beta)+(1-\pi)\alpha}$ & $1-\log \pi$ \\
        Gain Curve &  percentage: $\pi (1-\beta)+(1-\pi)\alpha$ & recall: $1-\beta$ & $1-\frac{\pi}{2}$ \\
    \hline
    \end{tabular}
    \caption{Formulas of some scoring metrics terms of $\{n, \pi, \alpha, \beta\}$.} 
    \label{tab:scoring-metrics}
\end{table}
\footnotetext{The percentage of instances predicted as positive by the classifier.}

It is not trivial to find the analytic formula of the curve, i.e., the functional relationship between the y- and x-coordinates. In practice, these curves are approximately plotted by applying a sequence of $J$ thresholds $1=t_1>t_2>\ldots>t_{J-1}>t_J=0$ to obtain $J$ confusion matrices. For each confusion matrix, x- and y-coordinates are computed, and a curve is formed by connecting the resulting $J$ points. In short, a scoring metric $\mathbf{M}_S$ is approximated using a sequence of confusion matrices $C_1, C_2, ..., C_J$ obtained by decreasing the thresholds $t$ from 1 to 0. 

To summarize the curve into a single numeric value, the \textit{area under the curve} (AUC) is often calculated. For all curves, the larger the AUC the better the performance. The last column of Table~\ref{tab:scoring-metrics} shows the ideal AUC when the predicted labels perfectly match the true labels. We can divide the AUC to the ideal value to obtain the Normalized AUC or NAUC, whose values range from 0 to 1:
\begin{equation}
    \textit{NAUC} = \frac{AUC}{\text{ideal } AUC} = \frac{AUC}{AUC \text{ induced by the ideal classifier}}. \label{equ: NAUC}
\end{equation}
The NAUC measures how close the classifier of interest is to the ideal classifier that predicts everything correctly. 

\subsection{Issues with Confusion-Matrix-Based Performance Metrics} \label{sec:issues}

\subsubsection{Dependence on Imbalance Rates} 

From Tables~\ref{tab:labeling-metrics} and \ref{tab:scoring-metrics}, we can see that some metrics, such as ROC, depend only on the classification errors $\{\alpha, \beta\}$, and thus are robust to class imbalance. However, due to the loss of information about class distribution, these metrics may be less informative in imbalanced cases, such as fraud detection \cite{lobo2008auc}. Other metrics such as f1\_score, MCC, and PRC, whose formulas include the prevalence rate $\pi$, are preferable when the data is imbalanced \cite{chicco2021matthews,saito2015precision}. 

Nevertheless, metrics dependent on $\pi$ can be sensitive to class imbalance rates in the sense that $\pi$ can affect the \textit{calibrations} or \textit{range} of these metrics, making it difficult to measure the quality of a classifier: \textbf{depending on $\pi$, a low nominal value may not necessarily indicate a poor performance}. The example below illustrates the dependence of f1\_score on class imbalance.

\begin{example}[Relative performance of f1\_score under class imbalance] \label{ex:issue-f1}
    Recall from Table~\ref{tab:labeling-metrics} that the f1\_score is given by 
    \begin{equation}
    f1(\pi, \alpha, \beta) = \frac{2 \pi (1-\beta)}{\pi(2-\beta)+(1-\pi)\alpha}. \label{eq:f1}
    \end{equation}
    Consider the trivial positive classifier (TPC), which always predicts positive (label 1). For any test set, this classifier will have $\alpha=1$ and $\beta=0$ and thus obtain an f1\_score of $f1(\pi,1,0) = \frac{2 \pi}{1+\pi}$.
    When the test set is balanced, i.e., $\pi = 0.5$, an arbitrary classifier with an f1\_score of 0.6 will have a poor performance because it is even worse than the f1\_score of the TPC at $0.6667$. On the contrary, when $\pi = 0.1$, the value $0.6$ for f1\_score suggests a good performance when compared to that of the TPC at $0.1818$.
\end{example}

In general, a decrease in the nominal values of a $\pi$-dependent metric may not imply a performance degradation. Changes in $\pi$ can cause significant fluctuations in calibration, making the nominal values obtained from $\pi$-dependent metrics unreliable in detecting classification performance drifts. In practice, the prevalence rate $\pi$ of the test set, such as the COVID-19 infection rate, often changes over time. Hence, it is not a good idea to \textit{monitor} classification performance by directly comparing the nominal values of these metrics on different test sets of different time periods. Moreover, in some cases, $\pi$ can even change the range of the metric. As a result, it is unreasonable to \textit{average} nominal metric values over multiple classes or test sets of different imbalance rates. This, in turn, creates difficulty in problems such as multi-class classification or cross-validation.

\subsubsection{No Free Lunch} 

There is an ongoing discussion about what metrics should be used to evaluate the performance of classification models \citep{chicco2021matthews,hossin2015review}. In fact, different performance metrics may be preferred in different applications. More specifically, the costs of Type-I error (false positive or false alarm) and Type-II error (false negative or miss) can be weighed differently across different domains \cite{ahmadzadeh2022contingency}. For example, lift curve is preferred in product recommendation as Type-I error is to be minimized, while PRC is preferred in adverse event identification as Type II-error is considered more serious. 

Because different performance metrics have different ranges of attainable values, and moreover, these ranges may vary with the prevalence of the test set, the interpretation of nominal metric values is inherently context-dependent. As a result, the same nominal value may reflect different relative performance under different prevalence conditions, making direct comparison and consistent interpretation difficult. This lack of consistent calibration complicates the use of fixed reference levels for performance monitoring, model comparison, or retraining decisions. This contrasts with settings such as statistical hypothesis testing, where conventional reference levels of 0.05 for p-values provide consistent interpretation across studies. These challenges motivate the development of a standardization method that places performance metric values on a common, interpretable scale, enabling consistent assessment across metrics and test sets. %to answer which value is good, compare between values

%-------------------------------------
\section{Methods} \label{sec:method}
%-------------------------------------

\subsection{A Universal Standardization} \label{sec:outperformance}

To address the problems discussed above, we seek a \textit{universal standardization} method for classification performance metrics such that even when applied to different test sets (with possibly different imbalance rates): (i) a decrease in the standardized value suggests a performance degradation; and (ii) a common criterion (e.g., a fixed threshold) can be uniformly chosen to evaluate the classifier's performance based on the standardized score.

%\item its values can be meaningfully averaged across multiple classes or test sets, regardless of class distributions or imbalance rates.

We propose that such a universal standardization can be achieved by evaluating the relative performance of a classifier within a distribution of all possible performances given the prevalence rate $\pi$. To build intuition, consider classifiers as students and the test set as an exam. Just as raw test scores may drop with tougher exams, nominal performance metric values can decline with increased class imbalance. Thus, evaluating students solely based on nominal scores can be misleading. A better approach is to consider the rank of the classifier’s performance among possible performances, similar to evaluating a student’s percentile rank over their raw test score. This perspective offers a more consistent and robust evaluation, especially when exam difficulty (or class imbalance) varies. Mathematically, we define such a universal standardization as the \textit{outperformance standardization function} in Definition~\ref{def:outperformance}.

\begin{definition}[Outperformance Standardization Function] \label{def:outperformance} Consider using a performance metrics $\mathbf{M}$ to evaluate a classifier of interest $\hat{f}$ on a given test set $\mathcal{D}_{\textnormal{test}} = \{(\mathbf{x}_i, y_i)\}_{i=1}^n$ and obtain a performance score $\mu$. Without loss of generality, suppose a higher value of $\mathbf{M}$ indicates a better performance. The \textbf{outperformance standardization value (o-value)} of the performance $\mu$ can then be defined as:
\begin{equation}
    \textnormal{OPS}_{\mathbf{M}}(\mu;\pi) = \Pr\{\mathbf{M}<\mu \mid P=\pi\},
\end{equation}
where $P$ denotes the random variable version of the prevalence rate $\pi$ and $\Pr$ is a probability measure over a reference distribution of performances. 
%Specifically, for a labeling metric $\mathbf{M}_L$, 
%\[\textnormal{OPS}_{\mathbf{M}_L}(\mu;\pi) = \Pr\{\mathbf{M}_L(A, B, P)<\mu|P = \pi\},\]
%and for a scoring metric $\mathbf{M}_S$ estimated using $J$ confusion matrices,
%\[\textnormal{OPS}_{\widehat{\mathbf{M}}_S}(\mu;\pi) = \Pr\{\widehat{\mathbf{M}}_S(A_1, ..., A_J, B_1, ..., B_J, P)<\mu|P = \pi\},\]
%where the probabilities are taken with respect to an assumed joint distribution given $P$ of $(A, B)$ or $(A_1, ..., A_J, B_1, ..., B_J)$, respectively.
\end{definition}

Essentially, the outperformance standardization value (o-value) is the probability that the observed performance \textit{outperforms} a pool of performances given the prevalence rate. A higher o-value indicates that the observed performance exceeds a larger proportion of possible performances, and therefore reflects better classification performance. Furthermore, as a probability, o-values always lie within the fixed range $[0,1]$, regardless of which CMBCP metric the OPS function applies to. In other words, o-values are standardized and scale-free. Their fixed range, consistent calibration, and clear interpretation enable consistent evaluation, comparison, and monitoring of model performances, even when changes in $\pi$ alter the calibration of nominal metric values. This OPS standardization also permits meaningful averaging of performances across multiple classes with different prevalence rates and supports the establishment of fixed reference criteria for assessing model performance.

It is important to note that instead of replacing the original performance metrics, the o-value acts as a \textit{meta-score}, providing complementary information on the percentile ranking of the observed performance. Specifically, when reporting results, we state that classifier $\hat{f}$ achieves a value of $\mu$ under metric $\mathbf{M}$ with an o-value of $\text{OPS}_{\mathbf{M}}(\mu;\pi)$, indicating that $\hat{f}$ outperforms $\text{OPS}_{\mathbf{M}}(\mu;\pi) \times 100\%$ of attainable classifier performances under the same prevalence conditions. As discussed above, this percentile-based interpretation facilitates consistent evaluation, comparison, and monitoring across metrics and test sets. In this sense, the role of the o-value is analogous to that of a p-value for a test statistic: it supplements the original metric with additional interpretive context, rather than replacing the underlying measurement. Finally, the o-value has the following useful property. 

\begin{property}[Linear Invariance Property] \label{prop:linear}
If $\mathbf{M}_1$ and $\mathbf{M}_2$ are two CMBCP metrics and $\mathbf{M}_2 = a \mathbf{M}_1 + b$ for some constants $a > 0$ and $b$, then $\textnormal{OPS}_{\mathbf{M}_1}(\mu;\pi) = \textnormal{OPS}_{\mathbf{M}_2}(a\mu+b;\pi)$. 
\end{property}

A proof of this property is given in Lemma 2 of the Supplementary Material. The linear invariance property is useful because many existing performance metrics have linear relations.  For example, $\textnormal{lift}=\textnormal{precision}/\pi$. Since the linear invariance property also holds when $a$ or $b$ are functions of $\pi$ as $P = \pi$ is given, the o-value of a classifier based on either lift or precision will be the same, and we only need to calculate the o-value for one of the two metrics.

In Definition~\ref{def:outperformance}, the outperformance standardization function is taken with respect to a reference distribution of performances. Recall that in the $\{n, \pi, \alpha, \beta\}$ parametrization of confusion matrices, while $n$ and $\pi$ encode information about the test set, $\alpha$ and $\beta$ encode information about the performance of the classifier. Therefore, choosing the reference distribution of performances is equivalent to choosing the joint distribution of Type-I and Type-II errors. In the following sections, we describe how we choose such distributions for labeling and scoring metrics, respectively. 

%-------------------------------------
\subsection{Outperformance Standardization for Labeling Metrics}
\label{sec:outperf-labeling}
%-------------------------------------

%Since conditioning on $\pi$ implies conditioning on subtotals $n_1+n_3$ and $n_2 + n_4$, the properties of multinomial distribution dictate that subvectors $(n_1, n_3)$ and $(n_2, n_4)$ are independent. Together with Equation~\eqref{eq:alpha-beta-pi}, this implies that for fixed test size and given prevalence rate, Type-I and Type-II errors are independent. 

Note that labeling metrics are functions of a single confusion matrix. For independent instances $\{(y_i, \hat{y}_i)\}_{i=1}^n$, the cells of the confusion matrix follow a multinomial distribution. The conditional independence property of the multinomial distribution implies that, given the prevalence rate $\pi$ and test size, Type-I and Type-II errors, $\alpha$ and $\beta$, are independent. Therefore, to calculate the o-values of labeling metrics, we consider the joint distribution in which Type-I error $A$ and Type-II error $B$ independently follow the $\mathrm{Unif}[0,1]$ distribution. Here, $A$ and $B$ denote the random variable versions of the Type I and Type II errors, i.e., $\alpha$ and $\beta$, respectively. The independent uniform distributions give equal probabilities to all possible combinations of $\alpha$ and $\beta$, helping us fairly evaluate the classifier of interest. Given $P=\pi$ and assuming $A, B \sim \mathrm{Unif}[0,1]$ independently, the outperformance standardization (OPS) function can be written as:
\begin{equation}
    \textnormal{OPS}_{\mathbf{M}_L}(\mu;\pi) = \Pr\{\mathbf{M}_L(A, B, P)<\mu \mid P = \pi\} = \int_0^1\int_0^1 \mathbb{I}(\mathbf{M}_L(\pi, \alpha, \beta)<\mu) d\alpha d\beta,
\end{equation}
where $\mathbb{I}(\cdot)$ denote the indicator function, $\mathbf{M}_L$ denote a labeling metric, and $P$ denote the random variable version of the prevalence rate $\pi$. For some labeling metrics, the OPS function can be derived analytically, such as f1\_score, as illustrated below in Example~\ref{ex:ops fi}. For others, o-values can be numerically estimated using either the trapezoidal rule or Monte Carlo approximation. 

%We discussed in Section~\ref{sec:outperformance} that $n$ can limit possible values of Type-I and Type-II errors. Hence, by assuming $A$ and $B$ independently follow $\mathrm{Unif}[0,1]$ distribution, we implicitly assume infinite test size $n$. Thus, a careful interpretation of the OPS value (for labeling metrics) is "the probability that the classifier outperforms random (equally likely) performances, given that it gives a similar performance on an infinitely large test set with class imbalance rate $\pi$".

\begin{example}[OPS Function based on f1\_score] \label{ex:ops fi}
    Using the f1\_score formula given in Equation \eqref{eq:f1}, the OPS function for f1\_score can be computed as:
    \begin{equation}
    \textnormal{OPS}_{f1}(\mu;\pi) = \begin{cases}
    \frac{(1+\pi)\mu}{2\pi(2-\mu)}, & \text{if } 0 \le \mu \le \frac{2\pi}{1+\pi} \\
    \frac{(1+\pi)\mu}{2\pi(2-\mu)} - \frac{[(1+\pi)\mu-2\pi]^2}{2\pi(1-\pi)\mu(2-\mu)}, & \text{if } \frac{2\pi}{1+\pi} < \mu \le 1.
    \end{cases}
    \end{equation} 

    \begin{figure}[ht]
    \centering
      \includegraphics[width=1\linewidth]{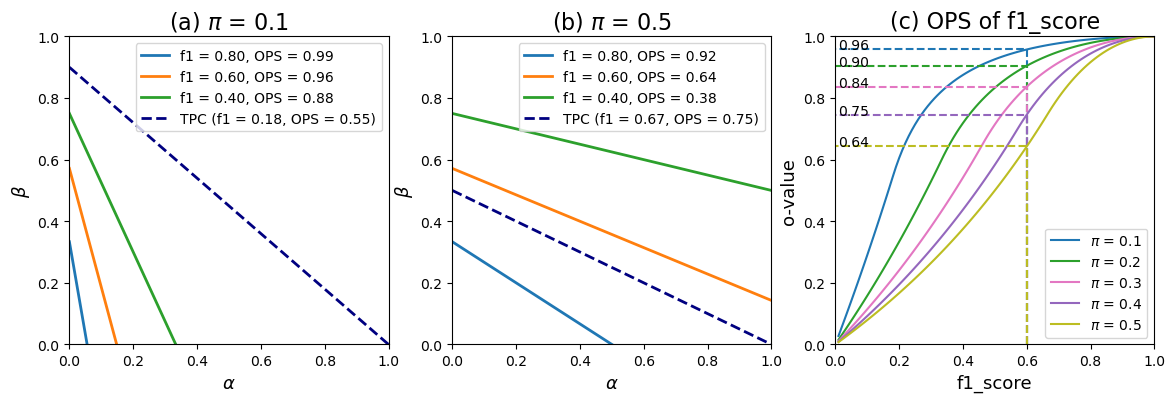}
      \caption{Geometric representation of $\textnormal{OPS}_{f1}$ when (a) $\pi=0.1$; (b) $\pi=0.5$. And (c) plots the OPS function of f1\_score given different $\pi$.}
    \label{cr_f1}
    \end{figure}

    The graphical representations of $\textnormal{OPS}_{f1}$ can be found in Figure \ref{cr_f1} (a) and (b). Given $P=\pi$, the performance space can be represented by a unit square $[0,1]^2$ that includes all possible pairs of $(\alpha, \beta)$. For a given $\mu \in (0,1)$, the straight line $f1(\pi, \alpha, \beta)=\mu$ divides the unit square into two parts, where the left part indicates better performance ($f1>\mu$), and right part indicates worse performance ($f1<\mu$). The o-value $\textnormal{OPS}_{f1}(\mu;\pi)$ is the area of the right part, representing the proportion of performances that are outperformed by $f1=\mu$. We can see that when $\pi = 0.1$, an f1\_score of 0.6 implies better performance than the TPC, while when $\pi = 0.5$, the same f1\_score of 0.6 implies worse performance than the TPC. This is consistent with what we discussed in Example~\ref{ex:issue-f1}. Figure \ref{cr_f1} (c) further illustrates how the prevalence rate $\pi$ changes the distribution of f1\_score. For the same value $f1=0.6$, the OPS function increases, and thus the performance of the classifier improves as $\pi$ decreases. As a result, lower values of the f1\_score could be satisfactory for highly imbalanced data while being inadequate for balanced data.
\end{example}

\subsection{Outperformance Standardization for Scoring Metrics}
\label{sec:outperf-scoring}

Scoring metrics $\mathbf{M}_S$ are usually \textit{estimated} by applying a sequence of thresholds $t_1, ..., t_J$ to the output scores $\{\hat{f}(\mathbf{x}_i)\}_{i=1}^n$ to obtain $J$ confusion matrices. Given $\pi$, these confusion matrices correspond to a sequence of Type-I errors $\alpha_1, ..., \alpha_J$ and a sequence of Type-II errors $\beta_1, ..., \beta_J$. Therefore, the OPS function of a scoring metric can be estimated by
\[\textnormal{OPS}_{\widehat{\mathbf{M}}_S}(\mu;\pi) = \Pr\{\widehat{\mathbf{M}}_S(A_1, ..., A_J, B_1, ..., B_J, P)<\mu \mid P = \pi\},\]
where $\Pr$ is taken with respect to a joint distribution of Type-I and Type-II errors $(A_1, ...., A_J, B_1, ..., B_J)$. As shown in Equation~\eqref{eq:thresholding}, if $t_j < t_k$, then the number of predicted positive labels using threshold $t_j$ will be greater than that using $t_k$, implying that if $A_j > A_k$, $B_j < B_k$. Therefore, when calculating the o-values for scoring metrics, we must choose a reference distribution of $(A_1, A_2, ..., A_J, B_1, B_2, ..., B_J)$ that preserves this opposite ordering property.

\begin{figure}[tbh!]
\centering
\begin{tikzpicture}[
roundnode/.style={circle, draw=red!60, fill=white!5, very thick, minimum size=8mm},
whitenode/.style={circle, draw=white!60, fill=white!5, very thick, minimum size=10mm},
ellipsenode/.style={ellipse, draw=red!60, fill=white!5, very thick, minimum height=6mm, minimum width=10mm},
rectanglenode/.style={rectangle, draw=red!60, fill=white!5, very thick, minimum height=8mm, minimum width=12mm, align=center},
]

%Nodes
\node[rectanglenode]   (c4)   at (-4.5, 3) {\small $\alpha_4 \in [0, \alpha_2)$ \\[-5pt] \small $\beta_4 \in (\beta_2, 1]$};
\node[rectanglenode]   (c5)   at (-1.5, 3) {\small $\alpha_5 \in (\alpha_2, \alpha_1)$ \\[-5pt] \small $\beta_5 \in (\beta_1, \beta_2)$};
\node[rectanglenode]   (c6)   at (1.5, 3) {\small $\alpha_6 \in (\alpha_1, \alpha_3)$ \\[-5pt] \small $\beta_6 \in (\beta_3, \beta_1)$};
\node[rectanglenode]   (c7)   at (4.5, 3) {\small $\alpha_7 \in (\alpha_3, 1]$ \\[-5pt] \small $\beta_7 \in [0, \beta_3)$};

\node[rectanglenode]   (c2)   at (-3, 4.5) {\small $\alpha_2 \in [0,\alpha_1)$ \\[-5pt] \small $\beta_2 \in (\beta_1, 1]$};
\node[rectanglenode]   (c3)   at (3, 4.5) {\small $\alpha_3 \in (\alpha_1, 1]$ \\[-5pt] \small $\beta_3 \in [0, \beta_1)$};

\node[rectanglenode]   (c1)   at (0, 6) {\small $\alpha_1 \in [0,1]$ \\[-5pt] \small $\beta_1 \in [0,1]$};

%Lines
\draw[thick,->] (c1) -- (c2);
\draw[thick,->] (c1) -- (c3);
\draw[thick,->] (c2) -- (c4);
\draw[thick,->] (c2) -- (c5);
\draw[thick,->] (c3) -- (c6);
\draw[thick,->] (c3) -- (c7);
\end{tikzpicture}
\caption{A Directed Binary Tree distribution with depth = 2.}
\label{fig:dbt}
\end{figure}
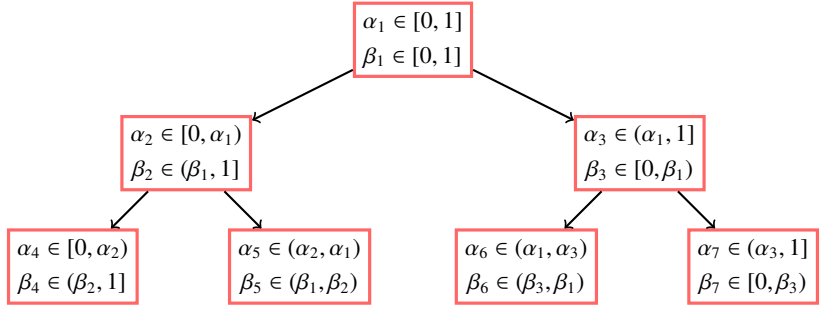

Specifying such a distribution where all possible performances are equally likely is challenging, and even if it existed, deriving analytic expressions for the OPS function would be difficult. Therefore, we estimate o-values using Monte Carlo approximation, and propose the \textit{Directed Binary Tree} (DBT) distribution, which is easy to sample from, as the joint distribution of $(A_1, A_2, ..., A_J, B_1, B_2, ..., B_J)$. In particular, we can draw a sample from the DBT distribution by first drawing the Type-I and Type-II errors $\alpha_1$ and $\beta_1$ independently from the $\textnormal{Unif}[0, 1]$ distribution. The obtained values $\alpha_1$ and $\beta_1$ each divide the original range $[0,1]$ into two parts. These parts can be used to draw the next two sets of errors: $(\alpha_2, \beta_2)$ drawn uniformly from the left part of $\alpha_1$, i.e., $\alpha_2 \sim \textnormal{Unif}[0,\alpha_1)$, and right part of $\beta_1$, i.e., $\beta_2 \sim \textnormal{Unif}(\beta_1, 1]$; and $(\alpha_3, \beta_3)$ can be drawn from the right part of $\alpha_1$, i.e., $\alpha_3 \sim \textnormal{Unif}(\alpha_1,1]$, and left part of $\beta_1$, i.e., $\beta_3 \sim \textnormal{Unif}[0,\beta_1)$. The next four sets of errors can be drawn similarly using $(\alpha_2, \beta_2)$ and $(\alpha_3, \beta_3)$, and so on. This procedure returns a sample $(\alpha_1, ..., \alpha_J, \beta_1, ..., \beta_J)$ that satisfy the opposite ordering we discussed above. We visualize the sampling procedure for the DBT distribution using a directed binary tree in Figure~\ref{fig:dbt}.

%let $E = (A, B)$ and $\theta = (l_A, u_A, l_B, u_B)$ where $(l_A, u_A)$ is the possible range of $A$ and $(l_B, u_B)$ is the possible range of $B$. W

%\begin{algorithm}
%\caption{Sampling from the Directed Binary Tree distribution}\label{algo:dbt}
%\SetKwInOut{Input}{Input}
%\SetKwInOut{Output}{Output}
%\Input{Number of Iteration $K$}
%\Output{A sample from the Directed Binary Tree distribution of $J = 2^{K-1}+1$ points}
%$ActiveRange \gets $ list that contains (0, 1, 0, 1)\;
%$NewActiveRange \gets $ empty list\;
%$SampleSet \gets $ empty list\;
%\For{$k \gets 1$ \KwTo $K$}{
%    \ForEach{$l \in$ $ActiveRange$}{
%    Sample $\alpha \sim \textnormal{Unif}[l_A, u_A]$ and $\beta \sim \textnormal{Unif}[l_B, u_B]$\;
%    Add $e = (\alpha, \beta)$ into $SampleSet$\;
%    Add $\theta_l = (l_A, \alpha, \beta, u_B)$ and $\theta_r = (\alpha, u_A, l_B, \beta)$ into $NewActiveRange$ \;
%    }
%    $ActiveRange \gets NewActiveRange$\;
%    $NewActiveRange \gets $ empty list\;
%}
%\Return $SampleSet$.
%\end{algorithm}

With $G$ samples from the binary tree distribution, we can estimate the o-value to be 
\begin{equation}
    \textnormal{OPS}_{\widehat{\mathbf{M}_L}}(\mu;\pi) \approx \frac{1}{G} \sum_g \mathbb{I}\left(\widehat{\mathbf{M}}_L(\alpha^{(g)}_1, ..., \alpha^{(g)}_J, \beta^{(g)}_1, ..., \beta^{(g)}_J, \pi)<\mu\right).
\end{equation}

Each DBT sample represents a possible performance curve from the reference distribution. We can compute the scoring metric based on this curve and compare it with the obtained value from the classifier of interest. By aggregating such comparisons over a large number of samples, we can estimate the o-value by Monte Carlo. In practice, DBT samples can be pre-generated and reused for efficiency.

There are two types of scoring metrics that are typically of interest: (i) the area under the curve, i.e., AUC; and (ii) a specific point $(u, v)$ on the curve that captures the trade-off between the two metrics in the x- and y-axes. For the second type, practitioners often fix a value on the x-axis (e.g., recall = 0.8 in a PRC) and evaluate the corresponding y-axis value (e.g., precision) on the curve.

Now, the $\text{AUC}^{(g)}$ corresponding to the $g$th DBT sample, given by $(\alpha^{(g)}_1, ..., \alpha^{(g)}_J, \beta^{(g)}_1, ..., \beta^{(g)}_J)$, can be approximated by linearly connecting points $(u^{(g)}_j, v^{(g)}_j)$ on the curve, which in turn corresponds to $(\alpha_j^{(g)}, \beta_j^{(g)})$. However, obtaining the coordinate $v$ for a specific $u$ from the sample $(\alpha^{(g)}_1, ..., \alpha^{(g)}_J, \beta^{(g)}_1, ..., \beta^{(g)}_J)$ is nontrivial, as there may be no index $k$ that exactly match the $g$th sample $(\alpha_k^{(g)},\beta_k^{(g)})$ to pre-specified  value $u$. To resolve this, we apply linear interpolation: identify neighbors  $u^{(g)}_l \le u \le u^{(g)}_r$ for some $l,r \in \{1,...,J\}$, and their associated errors $(\alpha_l^{(g)}, \beta_l^{(g)})$ and $(\alpha_r^{(g)}, \beta_r^{(g)})$, then estimate $(\hat{\alpha}, \hat{\beta})$ at $u$ via interpolation:
\begin{align}
    \hat{\alpha} = \alpha_l^{(g)} + (\alpha_r^{(g)}-\alpha_l^{(g)})\frac{u-u_l^{(g)}}{u_r^{(g)} - u_l^{(g)}}, \text{  and  } \hat{\beta} = \beta_l^{(g)} + (\beta_r^{(g)}-\beta_l^{(g)})\frac{u-u_l^{(g)}}{u_r^{(g)} - x_l^{(g)}}.
\end{align}
Finally, estimate $v$ from $\hat{\alpha}$ and $\hat{\beta}$ using the axis formula of the curve.

\begin{example}[OPS function of PRC] \label{ex:ops prc}
    One commonly used scoring metric to evaluate classification performance on imbalanced data is the PRC, which illustrates the trade-off between recall (x-axis) and precision (y-axis). 
    The OPS function for the AUC of the PRC is plotted in Figure~\ref{fig:OPS_AUC_PRC} for various prevalence rates $\pi$, ranging from 0.1 to 0.5. We can see how the distribution of AUC(PRC), therefore the o-values, change with respect to $\pi$. Specifically, for a given AUC value, such as 0.6, the OPS function increases as the data becomes more imbalanced. That is, when the data is imbalanced, a low value of AUC(PRC) may already imply a satisfactory performance: at $\pi = 0.1$, an AUC(PRC) of 0.6 implies that the classifier outperforms $96\%$ of classifiers in terms of AUC(PRC).
    \begin{figure}[tbh!]
    \centering
    \begin{subfigure}[b]{0.48\textwidth}
         \centering
         \includegraphics[width=\textwidth]{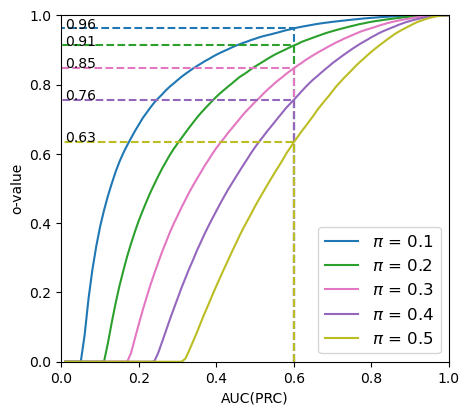}
         \caption{OPS of AUC(PRC)}
         \label{fig:OPS_AUC_PRC}
    \end{subfigure}
    \vspace{2mm}
    \begin{subfigure}[b]{0.48\textwidth}
         \centering
         \includegraphics[width=\textwidth]{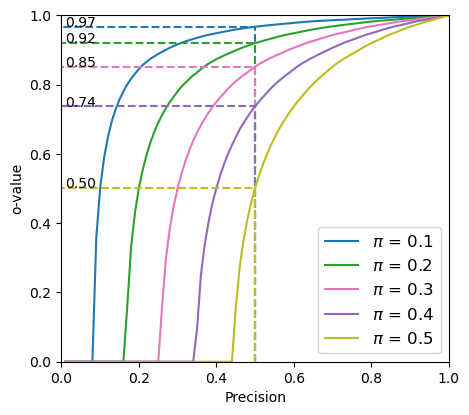}
         \caption{OPS of Precision $@$ Recall=0.8}
         \label{fig:OPS_Precision_Recall}
    \end{subfigure}
    \caption{The OPS functions for PRC with regard to: (a) AUC, and (b) the Precision at Recall=0.8, conditional on different imbalance rates.}
    \label{fig:OPS PRC}
    \end{figure}
    
    For a specific point $(u, v)$ on the PRC, we can compute the o-value of $precision=v$ given $recall=u$ using
    \begin{equation}
    \textnormal{OPS}_{precision|recall}(v,u;\pi) = \Pr\{precision<v \mid recall=u, P=\pi\}.
    \end{equation}
    As depicted in Figure~\ref{fig:OPS_Precision_Recall}, the OPS function of precision given recall = 0.8 increases as $\pi$ decreases. This indicates that, in imbalanced data, achieving a high recall does not necessarily require a high precision, and a lower precision may still reflect satisfactory performance. For example, when $\pi = 0.1$, at recall = 0.8, achieving precision = 0.5 already implies outperforming 97\% of classifiers.
\end{example}

In Examples~\ref{ex:ops fi} and \ref{ex:ops prc}, we have seen how class imbalance influences the distribution of nominal metric values, complicating performance evaluation. The OPS function standardizes these nominal values by mapping them to o-values with a direct percentile interpretation: an o-value of $x\%$ indicates that the observed performance exceeds $x\%$ of attainable classifier performances under the same CMBCP metric and prevalence conditions. Furthermore, since for a given metric, the OPS function uses the same reference distribution of performances, we can compare performances evaluated on different test sets. This is particularly useful for practitioners monitoring classification performances over time or evaluating performances in multi-label problems. 

%-------------------------------------
\section{Experiments}  \label{sec:experiment}
%-------------------------------------

In this section, we demonstrate how the proposed OPS method various confusion-matrix-based classification metrics for model performance monitoring across test sets with different prevalence rates. 

\subsection{Experimental Setup}
We use the Heart Disease Dataset \citep{Heart-Disease} and the Loan Default Dataset \citep{Loan-Default} for our experiments. The Heart Disease Dataset contains 253,680 records from the 2015 Behavioral Risk Factor Surveillance System, and the task is to predict heart disease using 21 features. The Loan Default Dataset includes 255,347 loan holders described by 17 personal and loan-related features, and the goal is to predict high-risk defaults. As shown in Table~\ref{tab:test set}, for each dataset, we construct three test sets with increasing prevalence:
\begin{itemize}
    \item \textbf{General set}: a random sample from the full dataset.
    \item \textbf{Cohort-based set}: a \emph{covariate-shifted} test scenario where test cases are restricted to individuals aged over 70 years (for the Heart Disease dataset) and with annual income under \$35{,}000 (for the Loan Default dataset), respectively. These test sets are designed to examine whether such a covariate shift leads to performance drifts on subgroups. 
    \item \textbf{Prevalence-control set}: a \emph{controlled prevalence-shift} scenario created by independently sampling from each class at a 3:7 ratio. Since this construction approximately preserves within-class feature distributions, the model's discriminative ability is not expected to change fundamentally. Any nominal metric changes may be driven by the prevalence shift and/or by the fixed decision threshold used for labeling metrics.
\end{itemize}
For each dataset, we train an XGBoost classifier \citep{chen2016XgBoost} on the remaining samples. The goal is to evaluate each XGBoost classifier on the corresponding three test sets.

\begin{table}[htbp!]
    \centering
    
    \begin{tabular}{|c|c|c|c|}
    \hline
    \multicolumn{4}{|c|}{\textbf{Heart Disease Dataset}}\\
    \hline
         & General Set & Elder Set & Prevalence-control Set  \\
    \hline
        Sample Size & 9,000 & 9,043 & 9,206 \\
        Prevalence & 0.09 & 0.19 & 0.30 \\
    \hline
    \multicolumn{4}{|c|}{\textbf{Loan Default Dataset}}\\
    \hline
     & General Set & Low-income Set & Prevalence-control Set  \\
    \hline
        Sample Size & 10,000 & 10,108 & 10,063 \\
        Prevalence & 0.11 & 0.20 & 0.30 \\
    \hline
\end{tabular}
    \caption{Test-set summaries. For each dataset, the General set is a random sample from the full dataset; the Elder (Heart Disease) and Low-income (Loan Default) sets are random samples restricted to individuals aged over 70 years and with annual income under \$35{,}000, respectively; and the Prevalence-control sets are sampled independently from each class at a 3{:}7 ratio.} 
    \label{tab:test set}
\end{table}

\subsection{Experimental Results}

We report (i) \textit{single-threshold} labeling metrics for prediction application, and (ii) \textit{curve-based} scoring metrics for risk identification application. For the single-threshold prediction evaluation, we treat Type-I and Type-II errors as equally important and therefore report f1\_score and MCC \cite{powers2011evaluation,chicco2021matthews}. Predicted labels are obtained using a fixed threshold $t=0.19$, selected to maximize f1\_score on a random validation sets. For risk identification application, the goal is to achieve high precision while maintaining a required coverage level (i.e., recall). We therefore use the precision-recall curve (PRC) and summarize it using AUC(PRC) and the Precision at Recall$=0.9$ (P@R=0.9). 

All nominal metric values and their corresponding o-values are reported in Table~\ref{tab:main-results}. The results show that the nominal values of commonly used $\pi$-dependent metrics may change substantially across test sets as prevalence varies, which can obscure whether performance differences reflect true model degradation or simply a shift in class balance. The proposed o-value addresses this issue by standardizing each nominal value to a prevalence-conditioned percentile scale.

For the Heart Disease dataset, the Prevalence-control set achieves much higher nominal f1\_score, MCC, and PRC-based metrics than the General set. However, their o-values are comparable (mostly around 0.85-0.91), indicating similarly strong standardized performance. In contrast, the Elder cohort consistently exhibits lower o-values (about 0.78-0.82), suggesting relatively weaker subgroup performance despite nominal values that may appear to be comparable with those of the other two test sets.

For the Loan Default dataset, nominal metric values increase with prevalence from the General to the Prevalence-control set. However, o-values separate this prevalence effect from changes in model behavior. Specifically, the PRC-based o-values are essentially unchanged across the three test sets (AUC(PRC): 0.81-0.83, and P@R=0.9: 0.78-0.81), indicating that the classifier's quality remains stable and no evidence of subgroup performance drift for the low-income cohort. In contrast, the o-value of the f1\_score drops from $0.83$ (General) to $0.74$ (prevalence-control), showing that the fixed operating threshold becomes less favorable in the prevalence-control set. In a monitoring pipeline, this pattern would typically suggest that the decision threshold should be revisited, rather than indicating a fundamental loss of discriminative ability.

\begin{table}[tbh!]
\centering

\begin{tabular}{|l|c|c|c|c|c|}
\hline
 \textbf{Test set} & $\boldsymbol{\pi}$ &
\textbf{F1} & \textbf{MCC} & \textbf{AUC(PRC)} & \textbf{P@R=0.9} \\
\hline
\multicolumn{6}{|c|}{\textbf{Heart Disease Dataset}}\\
\hline
General  & 0.09 & 0.41 (0.89) & 0.35 (0.87) & 0.35 (0.87) & 0.18 (0.90) \\
Elder    & 0.19 & 0.45 (0.80) & 0.30 (0.78) & 0.42 (0.80) & 0.26 (0.82) \\
Prevalence-control & 0.30 & 0.61 (0.85) & 0.47 (0.86) & 0.69 (0.91) & 0.50 (0.90) \\
\hline
\multicolumn{6}{|c|}{\textbf{Loan Default}} \\
\hline
General    & 0.11 & 0.36 (0.83) & 0.27 (0.80) & 0.32 (0.81) & 0.15 (0.78) \\
Low-income & 0.20 & 0.48 (0.81) & 0.32 (0.79) & 0.49 (0.84) & 0.28 (0.78) \\
Prevalence-control      & 0.30 & 0.51 (0.74) & 0.34 (0.78) & 0.58 (0.83) & 0.38 (0.81) \\
\hline
\end{tabular}

\caption{XGBoost performance across test sets. Each entry is reported as \textit{nominal metric (o-value)}.
Nominal metrics can vary substantially as prevalence changes, whereas o-value maps each nominal value to a prevalence-conditioned percentile, enabling comparison across test sets with different $\pi$.}
\label{tab:main-results}
\end{table}

\begin{figure}[tbh!]
\centering

% Row 1: Heart Disease
\begin{subfigure}[t]{0.48\textwidth}
  \centering
  \includegraphics[width=\textwidth]{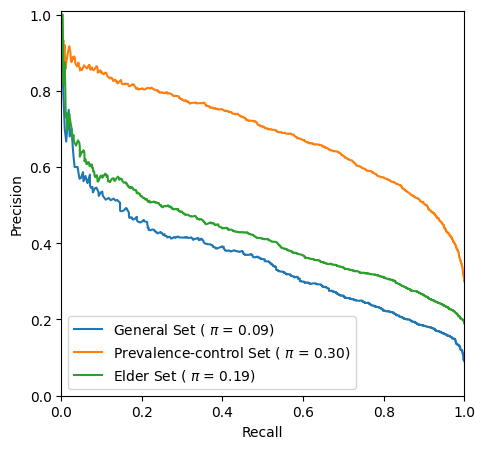}
  \caption{Heart Disease: PRC}
  \label{fig:prc-heart}
\end{subfigure}
\hfill
\begin{subfigure}[t]{0.48\textwidth}
  \centering
  \includegraphics[width=\textwidth]{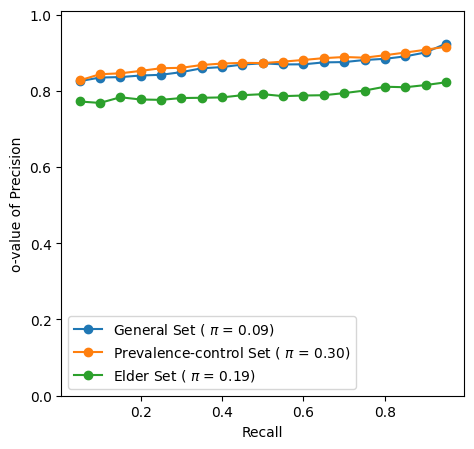}
  \caption{Heart Disease: OPRC}
  \label{fig:oprc-heart}
\end{subfigure}

\vspace{0.2cm}

% Row 2: Loan Default
\begin{subfigure}[t]{0.48\textwidth}
  \centering
  \includegraphics[width=\textwidth]{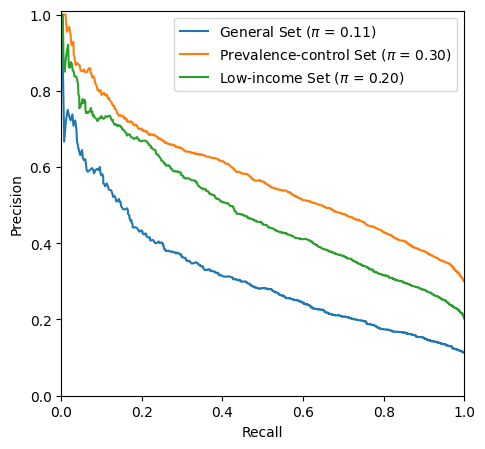}
  \caption{Loan Default: PRC}
  \label{fig:prc-loan}
\end{subfigure}
\hfill
\begin{subfigure}[t]{0.48\textwidth}
  \centering
  \includegraphics[width=\textwidth]{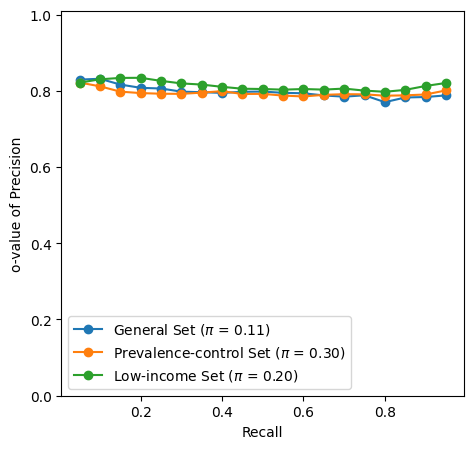}
  \caption{Loan Default: OPRC}
  \label{fig:oprc-loan}
\end{subfigure}

\caption{PRC (left) and OPS-standardized PRC (OPRC, right) for Heart Disease (top) and Loan Default (bottom).}
\label{fig:prc-oprc-2x2}
\end{figure}

In addition, to visualize OPS for curve-based evaluation, we introduce the \textit{OPS-standardized PRC (OPRC)}, which replaces the nominal precision on the y-axis with its corresponding o-value at each recall level on the x-axis (see Figure~\ref{fig:prc-oprc-2x2}). As shown in Figure~\ref{fig:prc-heart} and ~\ref{fig:prc-loan}, the PRCs exhibit a consistent pattern on both datasets: performances on test sets with higher prevalence tend to dominate those on test sets with lower prevalence. After applying OPS, as shown in Figure~\ref{fig:oprc-heart} and ~\ref{fig:oprc-loan}, the OPRCs are mapped to a common percentile scale, making cross-test-set comparisons clearer (e.g., weaker Elder-cohort performance for Heart Disease and nearly overlapping performance for Loan Default). Overall, OPRC provides a convenient visualization for monitoring PRC-based performance under prevalence shifts.

Additional lift-curve-based results for recommendation application are provided in Section~2 of the Supplementary Material.

\section{Conclusion} \label{sec:conclusion}

\subsection{Outperformance Standardization in Machine Learning Workflow}

In this paper, we introduced a novel standardization method for confusion-matrix-based classification performance (CMBCP) metrics, which we term the outperformance standardization (OPS) function. This method transforms commonly used performance metrics into a unified scale [0,1], enabling meaningful interpretation and comparison across varying contexts. The resulting o-values quantify the probability that a classifier's observed performance, given the prevalence rate, exceeds what would be expected by chance within a distribution of possible outcomes.

By applying the OPS function to a variety of classification performance metrics that are preferred in different applications, we demonstrated the generality and adaptability of our proposed method in the current applied machine learning workflow. Indeed, practitioners are not required to abandon their preferred metrics for their specific real-world application; instead, the OPS function provides additional information on percentile ranking that allows the values of those metrics to be better interpreted and assessed. More specifically, o-values can be incorporated into the reporting workflow in a similar fashion to how p-value is included in statistical reports for statistical significance. For example, a conclusion could be written as: the classifier attains an f1\_score of 0.6 (o-value: 96\%); and a simple interpretation can be given as: the classifier outperforms 96\% of attainable classifiers in terms of f1\_score. Furthermore, thanks to the common scale of [0,1] and the clear interpretation, the OPS function allows thresholds or standards to be established. For example, the 90\% threshold can be used to identify ``excellent" performance, while the 80\% threshold can be used to flag for re-training. Note that these thresholds are only examples, and specific thresholds or conventions for different applications in different domains need to be examined more carefully in future research.

Finally, the intuitive yet rigorous properties of the OPS function and o-values are particularly valuable in settings where class imbalance or dataset shifts affect the calibration of CMBCP metrics. This is especially important in real-world machine learning applications, where data distributions and imbalance rates often change over time, subgroups, or multiple labels, making reliable comparison and performance monitoring across test sets essential. Our experimental results, drawn from diverse real-world datasets and classification tasks, show that the proposed method is able to reveal and explain model degradation when covariate distribution changes, despite the seemingly unchanging nominal values of CMBCP metrics. This feature makes the OPS function a valuable tool not only for model selection and evaluation but also for detecting performance drifts over time or across different demographic or operational subgroups. Overall, the proposed outperformance standardization provides a flexible, interpretable, and statistically grounded approach to classifier evaluation - one that enhances transparency, comparability, and consistency in applied machine learning workflows.

\subsection{Related Works}

Recent work has explored principled representations and analyses of classification performance metrics. For example, \cite{luque2019impact} provides a systematic simulation-based analysis of how class imbalance affects CMBCP metrics, identifying clusters of metrics with similar behavior and advocating balance-aware evaluation measures such as Class Balance Metrics. Additionally, \cite{ahmadzadeh2022contingency} construct geometric representations of classifier performance based on joint distributions of true positive and true negative rates, providing structured spaces in which different metrics can be visualized and compared. These approaches primarily focus on characterizing relationships among metrics and understanding their geometric or structural properties. In contrast, our work addresses a complementary but distinct problem: the lack of a unified and interpretable standardization for assessing and comparing metric values. The proposed outperformance standardization does not redefine or re-weight classification performance criteria, but instead standardizes existing metrics by quantifying their relative standing among attainable performances under the same operating conditions.

Similarly, recent analyses such as the Worthiness Benchmark framework in \cite{shirdel2024worthiness} examine how metrics implicitly prioritize different confusion matrix components and characterize their ranking behavior. Such analyses provide insight into when particular metrics are appropriate, whereas the proposed outperformance standardization addresses the orthogonal challenge of making nominal metric values directly interpretable and comparable across metrics, test sets, and prevalence conditions. These perspectives are complementary: worthiness analysis clarifies what a metric measures, while the OPS function clarifies how strong a given performance is relative to its attainable range. Finally, prior work in \cite{hand2009measuring} demonstrates that even widely used metrics such as the AUC of ROC can suffer from fundamental interpretability limitations, showing that AUC(ROC) implicitly applies inconsistent misclassification cost trade-offs across classifiers and may therefore yield incoherent comparisons. This highlights the importance of principled and consistent performance interpretation. Our approach complements this line of work by providing a unified normalization framework that preserves the underlying evaluation criteria of existing metrics while enabling consistent interpretation and comparison of their values across classifiers, metrics, and test conditions.

Preliminary results of this work appeared as a short conference paper \cite{zhao2022classifier}. The present article is a substantially extended and revised version, adding detailed mathematical formulation and justifications, revisions and additional developments of methodology, detailed explanations, extensive experiments, and expanded discussions on utility, importance, limitations, related works, and future directions.

\subsection{Limitations and Future Directions}

A limitation of the proposed method is that it applies only to classification performance metrics defined as functions of the confusion matrix(ces). Consequently, it does not directly extend to performance measures based on probabilistic predictions or calibration, such as the Brier score, which are not solely determined by confusion matrix entries.

Furthermore, in this paper, we assumed that the Type-I and Type-II errors of the classifier of interest are measured correctly using the test set. In practice, a large test set should be used to ensure that the performance is measured correctly. More preferably, methods to quantify the uncertainty in measuring the Type-I and Type-II errors using finite test sets should be developed, so that confidence intervals for the o-values can be derived.

In addition, to calculate the o-values, we made specific choices on the reference distributions of performances due to their generality, ease in sampling, and uniform structure, which assign equal probabilities to attainable performances. In our opinion, these distribution features enable the o-values to be used across all applications, strengthening the universal application of our approach. However, it can also be possible that in specific domain applications, different weights can be given to different attainable performances. The construction, viability, and sensitivity analyses for alternative reference distributions are beyond the scope of this paper, and we leave them for future research.

Finally, because o-values are directly comparable across labels with varying imbalance rates, they are well-suited for evaluating multi-label classification performance. For instance, individual o-values can be computed for each label and then aggregated, either through simple averaging or other weighted schemes, to produce an overall score. Upcoming studies may explore optimal strategies for such aggregation. \\

\noindent \textbf{Acknowledgement}

\noindent This work was supported by Mitacs (grant number IT33843), Daesys
Inc., and Fin-ML. \\

\noindent \textbf{Declaration of generative AI and AI-assisted technologies in the writing process}

\noindent During the preparation of this work the author(s) used CHATGPT (Free version) in order to refine some of the language in the manuscript. After using this tool/service, the author(s) reviewed and edited the content as needed and take(s) full responsibility for the content of the publication.

%\appendix
%\section{Example Appendix Section}
%\label{app1}


%%       \cite{<label>} ==> [1]

%%
%% Example citation, See \cite{lamport94}.

\bibliographystyle{elsarticle-num} 
\bibliography{references}



%\begin{thebibliography}{00}

%% For numbered reference style

%\end{thebibliography}
\end{document}